\documentclass[letterpaper]{article} 
\usepackage{aaai2027}  
\usepackage[hyphens]{url}  
\usepackage{graphicx} 
\usepackage{natbib}  
\usepackage{caption} 
\usepackage{multirow}
\usepackage{amsmath,amssymb}
\usepackage{booktabs}
\usepackage[skins,breakable]{tcolorbox}
\usepackage{algorithm}
\usepackage{algpseudocode}
\usepackage[utf8]{inputenc}

\title{Style Wins, Substance Loses: A Diagnosis of LLM-as-Judge in Idea Generation}
\author{
Fengxian Ji\textsuperscript{\rm 1,2,4}\thanks{~~Equal contribution.},
Yuke Li\textsuperscript{\rm 1,3}\footnotemark[1],
Jingpu Yang\textsuperscript{\rm 4}\footnotemark[1],
Juanfan Wu\textsuperscript{\rm 1},
Fan Zhang\textsuperscript{\rm 2},
Zhexuan Cui\textsuperscript{\rm 6}\\
Yu Xie\textsuperscript{\rm 5},
Min Peng\textsuperscript{\rm 1},
Qianqian Xie\textsuperscript{1}\thanks{Corresponding author.},
Xiuying Chen\textsuperscript{\rm 2},
Zhuohan Xie\textsuperscript{2}\footnotemark[2] \\
}

\affiliations{
\renewcommand{\arraystretch}{1.15}
\begin{tabular}{c}
\textsuperscript{\rm 1}School of Artificial Intelligence, Wuhan University, \textsuperscript{\rm 2}MBZUAI ,\textsuperscript{\rm 3}Northeastern,\textsuperscript{\rm 4}Zhongguancun Academy
\\ \textsuperscript{\rm 5}The University of Hong Kong, \textsuperscript{\rm 6}Nanyang Technological University\end{tabular}
\\[2mm]
\texttt{\{fengxian.ji, fan.zhang, zhuohan.xie, xiuying.chen\}@mbzuai.ac.ae} \\
\texttt{llylykykk@gmail.com, vin019843@gmail.com, xiey@mails.neu.edu.cn}\\
\texttt{\{pengm, xieq\}@whu.edu.cn}, \texttt{jingpuyang290@gmail.com, juanfanwu@gmail.com}
}
\nocopyright

\begin{document}
\maketitle
\begin{abstract}
With the rapid growth of LLM-based scientific agents, scientific idea generation has become a key component of AI-driven research, highlighting the need for reliable LLM-as-Judge systems.
However, whether these judges truly evaluate the scientific substance of ideas or are influenced by superficial stylistic presentation remains an open question.
To address this question, we propose \textbf{\textit{SciStyleBench}}, a unified three component Benchmark for diagnosing and mitigating stylistic bias in LLM-based idea evaluation:
(i) First, \textbf{\textit{SciStyleStage}}, a three-stage evaluation environment that applies controlled stylistic perturbations to fixed scientific content across three settings no context, fixed-domain context, and open-domain retrieval context covering 600 scientific ideas and 15 style variants, with 9,000 evaluation instances per setting;
(ii) Second, \textbf{\textit{SciStyleMetrics}}, a set of quantitative measures including Style Bias Index (SBI), Substance Recognition Rate (SRR), and Adversarial Win Rate (AWR) to characterize how stylistic variation affects scoring stability, substance discrimination, and ranking robustness;
(iii) Third, \textbf{\textit{SciStyleExtractor}}, a plug-and-play evaluation module that separates presentation style from scientific content by predicting style type and deviation before style-conditioned evaluation, enabling us to assess whether style awareness reduces stylistic bias.
Experiments on SciStyleBench reveal that direct LLM judges are sensitive to writing style and weak at substance discrimination, while SciStyleExtractor improves robustness by reducing SBI from $0.566$ to $0.501$ and increasing SRR/AWR from $0.504/0.554$ to $0.759/0.899$.
These results suggest that robust idea evaluation requires invariance to stylistic variation without sacrificing sensitivity to scientific substance. 
Overall, SciStyleBench provides a systematic framework for identifying, quantifying, and mitigating stylistic bias in scientific idea evaluation.

\end{abstract}

\section{Introduction}
\begin{figure}[t]
    \centering
\includegraphics[width=\columnwidth]{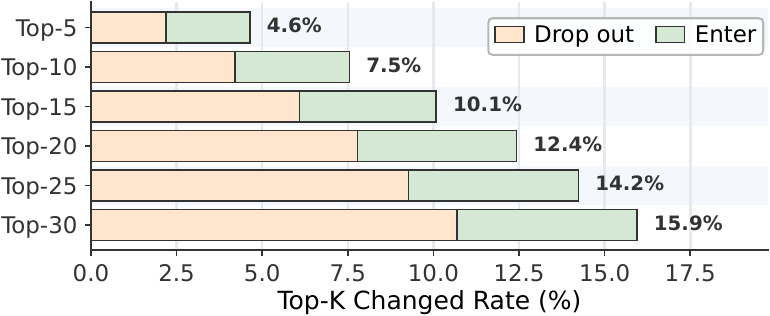}
\caption{
Style-induced changes in Top-K membership relative to the plain setting. Drop out and Enter denote leaving and entering the plain Top-K after transformation.}
\label{fig:topk-membership}
\end{figure}

With the rapid development of AI Scientist systems and automated research agents, scientific idea generation is gradually shifting from a one-off writing assistance task to a key generative component in automated research workflows \cite{lu2024aiscientistfullyautomated,Gottweis_2026,luo2025llm4srsurveylargelanguage}. 
Consequently, how to reliably evaluate and select from large pools of candidate ideas has become a central problem shaping the quality of downstream research processes \cite{guo2024ideabenchbenchmarkinglargelanguage,qiu2025aiideabench2025}.
In such systems, LLMs can continuously generate large numbers of candidate scientific ideas at very low cost\cite{si2024llmsgeneratenovelresearch,wang2024scipip,ji2026finestate}. 
However, downstream experimental validation, literature search, method implementation, paper writing, and human review all require substantial real-world resources ~\cite{gelles2024resourcedemocratization,si2025ideation,ye2026proof}. 
As a result, not every generated idea can be further executed~\cite{jie2026capability,liu2026researchbench,ji2026servimage,yang2026labguard,ji2026parametric}.
As shown in Fig.~\ref{fig:topk-membership}, stylistic transformations can
change Top-$K$ membership, causing ideas to enter or leave the selected set
despite retaining the same underlying scientific substance.
Therefore, idea evaluation is no longer merely an auxiliary step after
generation; it is a critical filtering mechanism that determines which ideas receive downstream research resources.

Existing research on the evaluation of AI-generated scientific ideas has developed multiple technical directions, aiming to improve evaluation reliability from the perspectives of scoring format, evaluation dimensions, information sources, and human calibration.
Representative approaches include ELO-based tournament-style pairwise ranking\cite{zheng2023judging,li2025automated,rabeyah2024llms},
multi-dimensional direct scoring along dimensions such as novelty, feasibility, and effectiveness
\cite{qiu2025aiideabench2025},
multi-agent specialized evaluation that incorporates external literature or specialized agents\cite{xiong2024kgcoi,liu2025personaflow,pu2025ideasynth,wang2024scimon}, and human-in-the-loop validation supported by expert blind review or human verification\cite{si2024llmsgeneratenovelresearch,radensky2026humanllmcompoundscientificideation,afzal2026beyond}.
However, despite these engineering improvements, most of these methods share the same structural premise: scalable idea evaluation still primarily relies on LLM-as-Judge.
Behind this premise, a more fundamental question remains insufficiently answered~\cite{wang2024large,sinhahajari2026limits}: do the scores produced by an LLM judge truly reflect the substantive value of a scientific idea, or are they shaped by surface-level effects introduced by its linguistic presentation style?

\begin{figure*}[htbp]
\centering
\includegraphics[width=\textwidth]{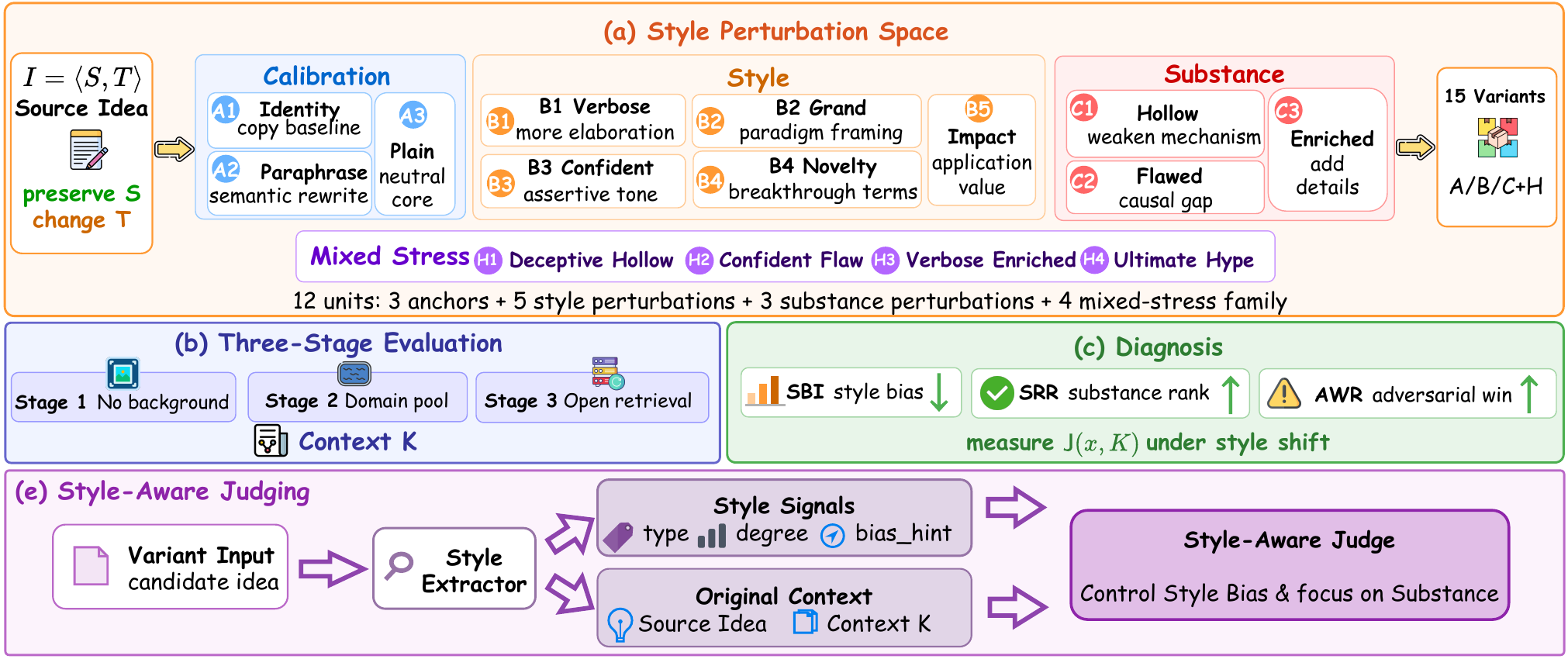}
\caption{
Overview of the SciStyleBench framework.
(a) Variant creation, (b) three-stage evaluation, (c) idea-variant taxonomy, (d) bias diagnosis with SBI, SRR, and AWR, and (e) style-aware judging with extracted style signals.
}
\label{fig:SciStyleBench}
\end{figure*}
To answer this question, it is not sufficient to simply observe whether an LLM judge’s scores change. What is needed is a complete evaluation framework that can measure, diagnose, and mitigate stylistic bias.
However, existing scientific idea evaluation still has clear gaps at three levels: metrics, benchmarks, and judges. 
First, at the benchmark level, existing data typically entangle content with style and lack a diagnostic environment for controlled style perturbation under fixed content. 
Meanwhile, these benchmarks lack reliable evaluation ground-truth construction, making it difficult to effectively disentangle stylistic factors from scientific merit\cite{chen2026mlr,kon2025exp,liu2026researchbench}. 
Second, at the metrics level, existing evaluations lack a formal metric system for quantifying stylistic bias, making it difficult to characterize how style affects judge scores, dimension-level scores, overall scores, and top-K selection\cite{liu2025hypobench,kulkarni2025scientific}. 
Finally, at the judge level, most LLM-as-Judge methods directly evaluate ideas from raw text, causing scientific substance and presentation style to become entangled. Existing approaches, including prompting, multi-dimensional scoring, and retrieval augmentation, do not explicitly model stylistic nuisance factors, making it difficult for judges to distinguish the effects of scientific merit from rhetorical presentation\cite{paperzilla_rag_retrieval_2024}.

To address these issues, we develop SciStyleBench, a benchmark for systematically studying stylistic bias in scientific idea evaluation consisting of three core components:
First, \textbf{SciStyleStage}, a style-paired benchmark that establishes
relative ground truth by holding scientific substance fixed while varying
presentation style, covering 600 scientific ideas, 15 controlled variants,
and three background settings, with 9,000 instances per setting.
Second, \textbf{SciStyleMetrics}, characterizes how writing style affects the scores and ranking outcomes produced by LLM judges through three closely interrelated metrics SBI, ADR, and SRR which respectively measure stylistic sensitivity, signal loss at the aggregation layer, and the ability to recognize substantive content.
Finally, \textbf{SciStyleExtractor} is a plug-and-play auxiliary judging module that
identifies style-related nuisance signals and injects structured information about style type, deviation from neutral presentation, and bias-control guidance into a frozen judge. 
Experiments on SciStyleBench show that Idea evaluators remain sensitive to style and weak in substance discrimination, while SciStyleExtractor improves SBI/SRR/AWR from $0.566/0.504/0.554$ to $0.501/0.759/0.899$. 
These results show that most existing idea evaluators are influenced by writing style rather than relying solely on substantive content. Improving their ability to assess ideas based on substance is therefore essential. SciStyleBench provides a systematic framework for identifying, quantifying, and mitigating such stylistic biases.

The main contributions of this paper are threefold.
(1) First, SciStyleBench, a style-paired benchmark with controlled presentation perturbations that disentangles style and substance effects and reveals style-induced evaluation biases.
(2) Second, SciStyleMetrics to quantify stylistic bias in scientific idea evaluation through metrics such as SBI, ADR, and SRR, with theoretical analysis of aggregation dilution.
(3) Finally, SciStyleExtractor, a plug-and-play auxiliary module that extracts structured style signals and injects bias-control guidance into frozen LLM judges to improve robust scientific idea evaluation on SciStyleBench.

\section{Related Work}

\paragraph{LLM-as-Judge in Scientific Idea Generation.}
LLM-as-Judge has become a common approach for evaluating AI-generated
scientific ideas. Existing methods typically use pairwise ranking or
multi-dimensional scoring based on criteria such as novelty, feasibility,
effectiveness, and falsifiability
\cite{shahhosseini2025large,bao2026contemporary}.
Recent benchmarks further evaluate whether LLM judges can identify
methodologically sound proposals or make temporally grounded judgments
\cite{ho2026soundnessbench,ye2026proof,wang2026firstresearch}. Other studies improve idea
evaluation through literature retrieval, specialized agents, multi-agent
review, and human calibration
\cite{baek2025researchagent,
li2026graph2idearetrievalaugmentedscientificideageneration,
keya2025sciideacontextawarescientificideation,
dong2026evolvingideagraphslearnable,
su2025headsbetteroneimproved,
radensky2024scideator,liu2026whoowns,nigam2024accelerontool}.
However, these approaches generally treat the linguistic realization of an
idea as fixed and do not test whether the same scientific substance receives
different evaluations under alternative presentation styles.

\paragraph{Stylistic Bias in LLM-as-Judge.}
Prior work shows that LLM judges are influenced by position, length,
verbosity, formatting, model identity, and other surface-level cues
\cite{li2024llms,shen2026navigating,koo2026auditing,ruan2026evaluatingllmsdivergentthinking}.
Benchmarks such as LLMBar test whether judges resist superficially
convincing responses, while subsequent studies examine stylistic preferences,
systematize evaluation biases, and propose mitigation methods
\cite{zeng2024evaluating,wu2025style,zhou2024mitigating,
zhou2026toward,yang2026any}. These findings suggest that judge outputs may
reflect presentation artifacts rather than task-relevant quality
\cite{cao2025evaluating,rasheed2026fluent}.
However, existing studies mainly focus on
general response evaluation and have not systematically examined
content-preserving style interventions for complete scientific ideas or their
effects on dimension-level scores, rankings, and Top-$K$ selection across
evidence settings. We address this gap with SciStyleStage,
SciStyleMetrics, and SciStyleExtractor.

\section{SciStyleStage}
\label{sec:scistylebench}
\subsection{Problem Formulation}
To investigates whether LLM-based judges remain stable when the
presentation style of a scientific idea changes while its scientific
substance is preserved. We represent an idea as
$I=\langle S,T\rangle$, where $S$ denotes its scientific substance,
including the research problem, methodological mechanism, variable
relationships, and verifiable contributions, while $T$ denotes its
presentation style, including verbosity, rhetorical strength, narrative
structure, confidence, and novelty framing. Given background information
$K$, the judge's evaluation is denoted by $J(I,K)$. Under fixed background knowledge $K$, different presentations of the same
scientific substance $S$ should receive similar evaluations. We test this
invariance through controlled style-only perturbations.

Our core intervention is a \textit{style-only perturbation}. A
perturbation operation $p$ transforms the original idea into
$I_p=p(I)=\langle S,T_p\rangle$, where $T_p\neq T$. Thus, $I$ and $I_p$
preserve the same scientific substance $S$ but differ in presentation
style. We define the resulting evaluation shift as
\begin{equation}
    \Delta_p(I,K)
    =
    J(I_p,K)-J(I,K).
    \label{eq:style-evaluation-shift}
\end{equation}

For a style-invariant judge, $\Delta_p(I,K)$ should remain close to zero
because the original idea and its style variant contain the same
scientific substance. A systematic positive or negative shift indicates
that the evaluation is influenced by presentation style rather than
scientific content, which we refer to as \textit{style-induced bias}.

\subsection{Style Perturbation Space}
\label{sec:variant-design}
We first selected 600 papers from NeurIPS and ICLR across several research domains, including Computer Science, Biology, Mathematics, Astronomy/Space Science, and Economics. We then used DeepSeek-V3 to generate style variants for each idea and validated their quality.
As shown in Fig.~\ref{fig:SciStyleBench} (a),
We define the style perturbation space using two types of variants: single-style and mixed-style variants. The single-style space contains 11 perturbations across three categories. Category A includes three baseline and expression-control variants: A1 (\textit{Identity}), which retains the original text; A2 (\textit{Paraphrase Only}), which rephrases the text while preserving its meaning; and A3 (\textit{Plain Core}), which removes rhetorical packaging. Category B includes five core \textit{style-only perturbations} that preserve the scientific substance while modifying only the presentation style: B1 (\textit{Verbose}), which expands the description with additional detail; B2 (\textit{Grand Narrative}), which introduces broad and visionary framing; B3 (\textit{Overconfident}), which expresses claims with greater certainty; B4 (\textit{Novelty Emphasis}), which strengthens novelty-related wording; and B5 (\textit{Application Framing}), which emphasizes practical value and potential impact. Category C includes three substance and logic contrast variants: C1 (\textit{Hollow}), which removes substantive support while retaining persuasive presentation; C2 (\textit{Flawed}), which introduces logical weaknesses; and C3 (\textit{Enriched}), which adds substantive information. The mixed-style space contains four variants: H1 (\textit{Deceptive Hollow}: B2 + B4 + C1), which combines hollow content with visionary and novelty framing; H2 (\textit{Confident Flaw}: B3 + C2), which presents flawed reasoning with strong confidence; H3 (\textit{Verbose Enriched}: B1 + C3), which combines detailed expression with substantive enrichment; and H4 (\textit{Ultimate Hype}: B2 + B4 + B5), which jointly emphasizes vision, novelty, and application value.

\subsection{Three-Stage Evaluation Tasks}
\label{sec:style-space}
As shown in Fig.~\ref{fig:SciStyleBench} (b),
SciStyleBench evaluates the same set of idea variants using identical judging prompts and scoring criteria under three evaluation settings that differ only in the literature context. Stage 1 provides no external literature, Stage 2 provides literature from a fixed domain-specific pool, and Stage 3 provides literature retrieved specifically for the source idea. This controlled design allows us to examine whether external evidence mitigates or amplifies the judge's sensitivity to presentation style.

\paragraph{Stage 1: No Background.}
The judge evaluates each idea variant without access to any external literature, relying solely on the information contained in the idea text.

\paragraph{Stage 2: Fixed-Domain Background.}
The judge receives literature from a fixed pool corresponding to the scientific domain of the source idea. These domain-specific literature pools are constructed from ResearchBench~\cite{liu2026researchbench}, Paperzilla 250~\cite{paperzilla_rag_retrieval_2024}, and IdeaBench~\cite{guo2024ideabenchbenchmarkinglargelanguage}. ResearchBench supplies most scientific domains, Paperzilla 250 supplies Computer Science, and IdeaBench supplies Medicine.

\paragraph{Stage 3: Idea-Specific Retrieval.}
We use the original, unperturbed source idea to construct retrieval queries covering its research question, methodology, and key concepts. Candidate papers are retrieved from OpenAlex, with Crossref used as a supplementary source when the OpenAlex results are insufficient. After deduplication and relevance ranking, we retain the 50 papers most relevant to each source idea.

In Stages 2 and 3, the judge does not receive the full text of the selected papers. Instead, each paper is represented by its title and abstract, formatted as a structured list and appended to the idea variant as background context. All variants derived from the same source idea receive the same literature in the same order, preventing retrieval differences from introducing an additional confounding factor.

\section{SciStyleMetrics}
\label{sec:scistylemetrics}

As shown in Fig.~\ref{fig:SciStyleBench} (c), SciStyleMetrics evaluates whether an LLM judge is insensitive to stylistic
variation while remaining sensitive to scientific substance. Let
$J^{(d)}(I_{i,p},K_i^{(m)})$ denote the score of idea variant $I_{i,p}$ on
dimension $d$ under evaluation setting $m$. We measure three complementary and interdependent properties using SBI, SRR, and AWR.

\paragraph{Style Bias Index.}
The Style Bias Index (SBI) measures whether the judge remains stable when the
scientific substance is fixed but the presentation style changes. Using the
Plain variant as the neutral reference and Paraphrase as a rewriting control,
we define
\begin{equation}
\resizebox{0.9\columnwidth}{!}{%
$\begin{aligned}
\mathrm{SBI}^{(m)}(p)
=
\mathbb{E}_{i,d}\Bigg[
\Bigg|
&
\left(
J^{(d)}(I_{i,p},K_i^{(m)})
-
J^{(d)}(I_{i,\mathrm{plain}},K_i^{(m)})
\right) \\
&-
\left(
J^{(d)}(I_{i,\mathrm{para}},K_i^{(m)})
-
J^{(d)}(I_{i,\mathrm{plain}},K_i^{(m)})
\right)
\Bigg|
\Bigg]
\end{aligned}$%
}
\end{equation}
A lower SBI indicates that different expressions of the same scientific idea
receive more consistent scores. Thus, SBI answers: \textit{when substance is
unchanged, how much does the score change because of style?}

\paragraph{Substance Recognition Rate.}
The Substance Recognition Rate (SRR) measures whether the judge detects
controlled changes in scientific quality. Let $\mathcal{Q}_{\mathrm{sub}}$
contain ordered pairs $(a,b)$ in which $a$ has stronger substance than $b$,
including Enriched over Plain, Plain over Hollow, and Plain over Flawed:
\begin{equation}
\resizebox{0.9\columnwidth}{!}{%
$\displaystyle
\mathrm{SRR}^{(m)}
=
\frac{1}{|\mathcal{Q}_{\mathrm{sub}}|}
\sum_{(a,b)\in\mathcal{Q}_{\mathrm{sub}}}
\mathbb{I}\left[
J(a,K^{(m)}) > J(b,K^{(m)})
\right].
$%
}
\end{equation}
A higher SRR indicates stronger sensitivity to substantive information,
methodological validity, and logical soundness. Thus, SRR answers:
\textit{can the judge recognize meaningful differences in scientific
substance?}

\paragraph{Adversarial Win Rate.}
The Adversarial Win Rate (AWR) measures which signal dominates when style and
substance conflict. Let $\mathcal{Q}_{\mathrm{adv}}$ contain pairs $(h,l)$,
where $h$ is a high-substance idea with plain presentation and $l$ is a
low-substance idea with persuasive stylistic packaging:
\begin{equation}
\resizebox{0.9\columnwidth}{!}{%
$\displaystyle
\mathrm{AWR}^{(m)}
=
\frac{1}{|\mathcal{Q}_{\mathrm{adv}}|}
\sum_{(h,l)\in\mathcal{Q}_{\mathrm{adv}}}
\mathbb{I}\left[
J(h,K^{(m)}) > J(l,K^{(m)})
\right].
$%
}
\end{equation}
A higher AWR indicates that the judge prioritizes scientific substance over
verbosity, grand narratives, novelty claims, or overconfident language. Thus,
AWR answers: \textit{when substance and presentation disagree, does the judge
select the substantively stronger idea?}

\paragraph{Joint Interpretation.}
The three metrics must be interpreted jointly: robust evaluation requires low
SBI, high SRR, and high AWR, corresponding to style invariance, substance
sensitivity, and adversarial robustness. In particular, low style sensitivity does not necessarily imply strong
evaluation ability. A judge that assigns nearly identical scores to all ideas
may obtain a very low SBI, while still failing to distinguish Enriched from
Plain, Plain from Hollow, or valid mechanisms from logically flawed ones.
Therefore, minimizing SBI alone may reward score collapse rather than genuine
robustness; it must be accompanied by high SRR and AWR. This distinction captures the central measurement principle of
SciStyleMetrics: robust evaluation is not simple score invariance, but
invariance to task-irrelevant nuisance variation while preserving sensitivity
to the target scientific construct.

\begin{figure}[t]
    \centering
\includegraphics[width=\columnwidth]{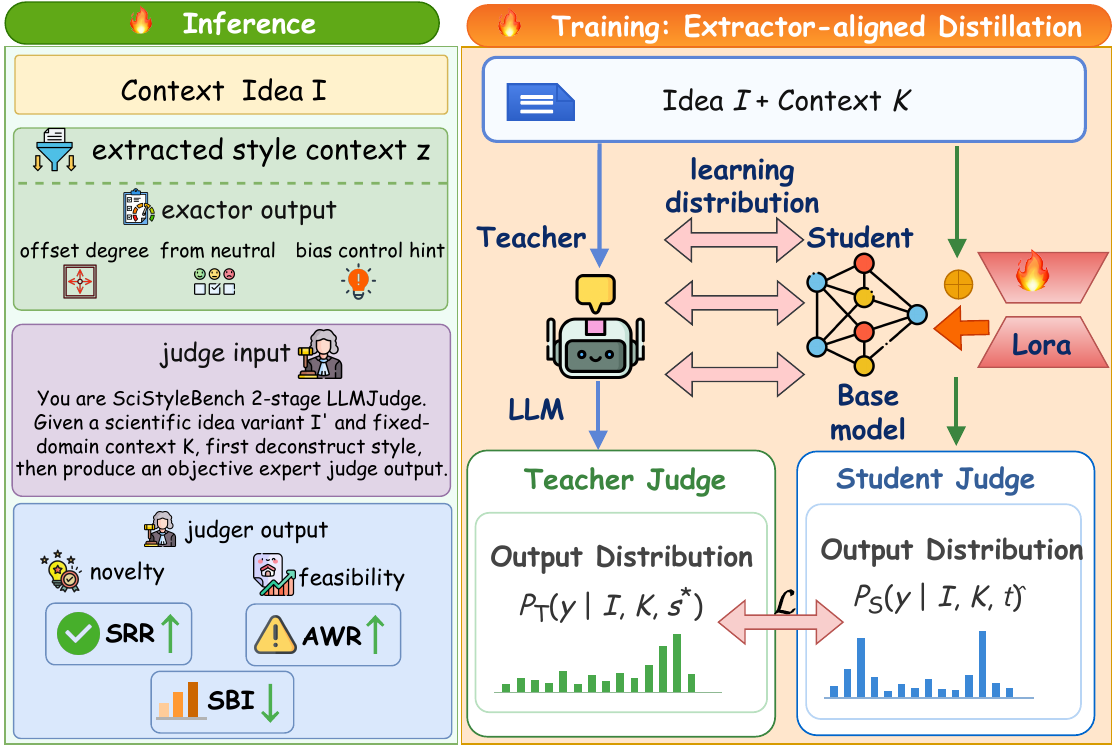}
\caption{
Training and inference pipeline of SciStyleExtractor.
The extractor learns teacher-aligned style signals during training and injects them into the judger at inference.}
\label{fig:scistyleextractor_pipeline}
\end{figure}

\section{SciStyleExtractor}
\label{sec:SciStyleExtractor}

Scientific substance and presentation style are entangled in the raw idea
text, making it difficult for an LLM judge to determine whether an apparent
quality signal comes from the idea itself or from rhetorical packaging.
SciStyleExtractor is therefore designed as an auxiliary judging module. It
does not score or rewrite the idea; instead, it identifies style-related
nuisance signals and injects structured bias-control information into a
frozen judge.

\subsection{Style-Aware Judging Framework}
\label{sec:SciStyleExtractor-framework}

Given an idea variant $I_p=\langle S,T_p\rangle$ and background context $K$,
a standard judge directly produces
$\mathbf{y}^{\mathrm{base}}_p=J_\theta(I_p,K)$ and may mistake stylistic
presentation for scientific quality. As shown in Fig.~\ref{fig:scistyleextractor_pipeline}, SciStyleExtractor introduces an
intermediate style signal before judging:

\begin{equation}
\mathbf{z}_p
=
f_\phi(I_p,K)
=
(t_p,\delta_p,h_p),
\hat{\mathbf{y}}_p
=
J_\theta(I_p,K;\mathbf{z}_p).
\end{equation}

Here, $f_\phi$ is the trainable style extractor and $J_\theta$ is the frozen
judge. The extractor output $\mathbf{z}_p$ contains the detected style type
$t_p$, its deviation from neutral presentation $\delta_p$, and a
bias-control instruction $h_p$. For example, it may identify an input as
verbose, determine that the change is stylistic rather than substantive, and
instruct the judge not to reward rhetorical elaboration. The final output
$\hat{\mathbf{y}}_p$ contains the judge's dimension-level scores, overall
score, and evaluation feedback. Thus, $\mathbf{z}_p$ serves only as auxiliary
evidence and does not directly determine the evaluation result.

\subsection{Extractor-Aligned Training}
\label{sec:SciStyleExtractor-training}

We train the extractor through teacher--student distillation. For each input
$(I_p,K)$, the teacher judge receives an ideal style signal
$\mathbf{z}_p^\star$, whereas the student judge receives the predicted signal
$\mathbf{z}_p=f_\phi(I_p,K)$. Their output distributions are denoted by
$P_{\mathrm{T}}(\mathbf{y})
=
P_\theta(\mathbf{y}\mid I_p,K;\mathbf{z}_p^\star)$ and
$P_{\mathrm{S}}(\mathbf{y})
=
P_\theta(\mathbf{y}\mid I_p,K;\mathbf{z}_p)$, respectively. We optimize the
extractor using

\begin{equation}
\mathcal{L}_{\mathrm{KL}}
=
\mathbb{E}_{(I_p,K)\sim\mathcal{D}}
\left[
\mathrm{KL}
\left(
P_{\mathrm{T}}(\mathbf{y})
\,\|\, 
P_{\mathrm{S}}(\mathbf{y})
\right)
\right].
\end{equation}

Minimizing this objective teaches the extractor to generate auxiliary style
signals that make the student approximate the teacher's debiased evaluation
behavior. During inference, the trained extractor generates
$\mathbf{z}_p$, which is injected into the frozen judge to support
substance-focused evaluation.

\begin{table*}[t]
\centering
\small
\setlength{\tabcolsep}{3.5pt}
\renewcommand{\arraystretch}{0.45}
\resizebox{\textwidth}{!}{
\begin{tabular}{lccccccccc}
\hline
\textbf{Judge Method}
& \multicolumn{3}{c}{\textbf{No Background}}
& \multicolumn{3}{c}{\textbf{Fixed-domain Background}}
& \multicolumn{3}{c}{\textbf{Idea-specific Retrieval}} \\
\cline{2-10}
& \textbf{SBI $\downarrow$} & \textbf{SRR $\uparrow$} & \textbf{AWR $\uparrow$}
& \textbf{SBI $\downarrow$} & \textbf{SRR $\uparrow$} & \textbf{AWR $\uparrow$}
& \textbf{SBI $\downarrow$} & \textbf{SRR $\uparrow$} & \textbf{AWR $\uparrow$} \\
\hline
\multicolumn{10}{l}{\textbf{Without SciStyleExtractor}} \\
Direct Qwen3.5-4B
& $0.780_{0.001407}$ & $0.860_{0.001204}$ & $0.940_{0.000564}$ & $0.476_{0.001180}$ & $0.700_{0.002100}$ & $0.850_{0.001275}$ & $0.660_{0.001133}$ & $0.840_{0.001344}$ & $0.910_{0.000819}$ \\
Direct Llama-3.1-8B
& $1.354_{0.008407}$ & $0.020_{0.000196}$ & $0.010_{0.000099}$ & $1.418_{0.011908}$ & $0.300_{0.002100}$ & $0.090_{0.000819}$ & $0.165_{0.000426}$ & $0.110_{0.000979}$ & $0.250_{0.001875}$ \\
Direct Qwen3.5-27B
& $0.308_{0.000443}$ & $0.510_{0.002499}$ & $0.630_{0.002331}$ & $0.262_{0.000368}$ & $0.440_{0.002464}$ & $0.520_{0.002496}$ & $0.265_{0.000365}$ & $0.420_{0.002436}$ & $0.550_{0.002475}$ \\
Direct DeepSeek-V3.2
& $0.487_{0.000760}$ & $0.760_{0.001824}$ & $0.750_{0.001875}$ & $0.418_{0.000650}$ & $0.690_{0.002139}$ & $0.730_{0.001971}$ & $0.202_{0.000472}$ & $0.400_{0.002400}$ & $0.420_{0.002436}$ \\
Direct OpenReviewer
& $0.494_{0.001248}$ & $0.160_{0.001344}$ & $0.610_{0.002379}$ & $0.078_{0.001221}$ & $0.000_{0.000000}$ & $0.000_{0.000000}$ & $0.238_{0.000837}$ & $0.020_{0.000196}$ & $0.170_{0.001411}$ \\
Direct AI-Scientist-Llama3.1-8B
& $0.478_{0.000959}$ & $0.450_{0.002475}$ & $0.720_{0.002016}$ & $0.164_{0.000712}$ & $0.170_{0.001411}$ & $0.160_{0.001344}$ & $0.177_{0.000357}$ & $0.300_{0.002100}$ & $0.220_{0.001716}$ \\
\hline
Style-CoT Qwen3.5-4B
& $0.665_{0.001008}$ & $0.770_{0.001771}$ & $0.830_{0.001411}$ & $0.860_{0.001622}$ & $0.680_{0.002176}$ & $0.680_{0.002176}$ & $0.714_{0.001427}$ & $0.800_{0.001600}$ & $0.810_{0.001539}$ \\
Style-CoT Llama-3.1-8B
& $1.325_{0.012569}$ & $0.220_{0.001716}$ & $0.180_{0.001476}$ & $0.788_{0.008262}$ & $0.220_{0.001716}$ & $0.170_{0.001411}$ & $0.241_{0.000485}$ & $0.080_{0.000736}$ & $0.090_{0.000819}$ \\
Style-CoT Qwen3.5-27B
& $0.484_{0.000636}$ & $0.760_{0.001824}$ & $0.850_{0.001275}$ & $0.452_{0.000782}$ & $0.770_{0.001771}$ & $0.760_{0.001824}$ & $0.451_{0.000872}$ & $0.650_{0.002275}$ & $0.710_{0.002059}$ \\
Style-CoT DeepSeek-V3.2
& $0.257_{0.000676}$ & $0.480_{0.002496}$ & $0.530_{0.002491}$ & $0.363_{0.000711}$ & $0.580_{0.002436}$ & $0.520_{0.002496}$ & $0.375_{0.000500}$ & $0.600_{0.002400}$ & $0.720_{0.002016}$ \\
Style-CoT OpenReviewer
& $0.472_{0.001791}$ & $0.060_{0.000564}$ & $0.430_{0.002451}$ & $0.008_{0.000029}$ & $0.000_{0.000000}$ & $0.000_{0.000000}$ & $0.282_{0.000976}$ & $0.110_{0.000979}$ & $0.210_{0.001659}$ \\
Style-CoT AI-Scientist-Llama3.1-8B
& $0.515_{0.000558}$ & $0.100_{0.000900}$ & $0.450_{0.002475}$ & $0.328_{0.001430}$ & $0.280_{0.002016}$ & $0.160_{0.001344}$ & $0.409_{0.000543}$ & $0.430_{0.002451}$ & $0.160_{0.001344}$ \\
\hline
\multicolumn{10}{l}{\textbf{With SciStyleExtractor}} \\
Qwen3.5-4B + LLM Style Injection
& $1.331_{0.001081}$ & $0.570_{0.002451}$ & $1.000_{0.000000}$ & $1.275_{0.001876}$ & $0.530_{0.002491}$ & $1.000_{0.000000}$ & $1.232_{0.001215}$ & $0.390_{0.002379}$ & $1.000_{0.000000}$ \\
Llama-3.1-8B + LLM Style Injection
& $0.467_{0.002704}$ & $0.160_{0.001344}$ & $0.170_{0.001411}$ & $1.803_{0.006790}$ & $0.280_{0.002016}$ & $0.400_{0.002400}$ & $0.350_{0.000552}$ & $0.140_{0.001204}$ & $0.900_{0.000900}$ \\
Qwen3.5-27B + LLM Style Injection
& $0.814_{0.002072}$ & $0.700_{0.002100}$ & $0.990_{0.000099}$ & $0.760_{0.002146}$ & $0.560_{0.002464}$ & $0.830_{0.001411}$ & $0.726_{0.001662}$ & $0.920_{0.000736}$ & $0.950_{0.000475}$ \\
DeepSeek-V3.2 + LLM Style Injection
& $1.009_{0.003917}$ & $0.780_{0.001716}$ & $1.000_{0.000000}$ & $1.251_{0.003194}$ & $0.910_{0.000819}$ & $1.000_{0.000000}$ & $0.967_{0.003332}$ & $0.780_{0.001716}$ & $1.000_{0.000000}$ \\
OpenReviewer + LLM Style Injection
& $0.555_{0.001507}$ & $0.230_{0.001771}$ & $0.460_{0.002484}$ & $0.006_{0.000023}$ & $0.000_{0.000000}$ & $0.000_{0.000000}$ & $0.535_{0.001715}$ & $0.210_{0.001659}$ & $0.370_{0.002331}$ \\
AI-Scientist-Llama3.1-8B + LLM Style Injection
& $0.419_{0.000250}$ & $0.090_{0.000819}$ & $0.480_{0.002496}$ & $0.251_{0.001423}$ & $0.050_{0.000475}$ & $0.190_{0.001539}$ & $0.618_{0.000492}$ & $0.360_{0.002304}$ & $0.880_{0.001056}$ \\
\hline
Qwen3.5-4B + Trained Style Extractor
& $0.363_{0.000366}$ & $0.910_{0.000819}$ & $0.960_{0.000384}$ & $0.369_{0.000382}$ & $0.920_{0.000736}$ & $0.970_{0.000291}$ & $0.406_{0.000286}$ & $0.840_{0.001344}$ & $0.930_{0.000651}$ \\
Llama-3.1-8B + Trained Style Extractor
& $0.912_{0.002047}$ & $0.550_{0.002475}$ & $0.990_{0.000099}$ & $0.392_{0.000627}$ & $0.740_{0.001924}$ & $0.960_{0.000384}$ & $0.568_{0.000712}$ & $0.420_{0.002436}$ & $0.880_{0.001056}$ \\
Qwen3.5-27B + Trained Style Extractor
& $0.500_{0.000478}$ & $0.890_{0.000979}$ & $0.920_{0.000736}$ & $0.504_{0.000759}$ & $0.870_{0.001131}$ & $0.820_{0.001476}$ & $0.513_{0.000376}$ & $0.890_{0.000979}$ & $0.860_{0.001204}$ \\
DeepSeek-V3.2 + Trained Style Extractor
& $0.415_{0.000380}$ & $0.730_{0.001971}$ & $0.880_{0.001056}$ & $0.598_{0.000805}$ & $0.710_{0.002059}$ & $0.860_{0.001204}$ & $0.469_{0.000390}$ & $0.640_{0.002304}$ & $0.760_{0.001824}$ \\
OpenReviewer + Trained Style Extractor
& $0.245_{0.000348}$ & $0.370_{0.002331}$ & $0.780_{0.001716}$ & $0.008_{0.000012}$ & $0.020_{0.000196}$ & $0.030_{0.000291}$ & $0.280_{0.000334}$ & $0.560_{0.002464}$ & $0.740_{0.001924}$ \\
AI-Scientist-Llama3.1-8B + Trained Style Extractor
& $0.920_{0.000514}$ & $0.690_{0.002139}$ & $0.990_{0.000099}$ & $0.452_{0.000449}$ & $0.840_{0.001344}$ & $0.960_{0.000384}$ & $0.702_{0.000937}$ & $0.810_{0.001539}$ & $0.940_{0.000564}$ \\
\hline
\end{tabular}
}
\caption{
Robustness comparison of judging methods across background settings.
Each metric is computed from five evaluation dimensions (novelty, feasibility,
significance, rigor, and clarity). Values report the mean performance, with
subscripts denoting standard errors. Lower SBI and higher SRR/AWR indicate
better robustness.
}
\label{tab:method_comparison}
\end{table*}

\section{Experiments}
\subsection{Experimental Setup}

\paragraph{Dataset and evaluation.}
We conduct experiments on SciStyleBench, which contains 600 source scientific ideas, each paired with 15 controlled variants covering reference rewrites, style-only perturbations, substance and logical controls, and mixed-style perturbations. All variants are evaluated under three background settings: no background, fixed-domain background, and idea-specific retrieval background, resulting in 9,000 evaluation instances per setting and 27,000 instances in total. We evaluate judging robustness using SBI, SRR, and AWR, which jointly measure style invariance, substance sensitivity, and adversarial robustness.

\paragraph{Judging methods.}
We evaluate six judges: four general-purpose models Qwen3.5-4B, Llama-3.1-8B, Qwen3.5-27B, and DeepSeek-V3.2 and two scientific-review models, OpenReviewer~\cite{idahl2025openreviewer} and AI-Scientist-Llama3.1-8B~\cite{lu2024aiscientistfullyautomated}. OpenReviewer is obtained by supervised fine-tuning Llama-3.1-8B for scientific reviewing, enabling a controlled comparison between the original backbone and its domain-adapted counterpart. 
For each judge, we compare four configurations: direct judging, style-aware CoT prompting, LLM-based style-signal injection, and our trained SciStyleExtractor. Direct judging uses the raw idea, whereas style-aware CoT prompts the judge to consider stylistic influence before scoring. The two injection-based settings provide the frozen judge with structured auxiliary signals, including style type, deviation from neutral presentation, and a bias-control instruction, generated by either a strong LLM or SciStyleExtractor. All judges remain frozen, so performance differences reflect the effect of auxiliary style awareness rather than further judge adaptation.

\begin{figure*}[htbp]
\centering
\scalebox{1}{%
\includegraphics[width=\textwidth]{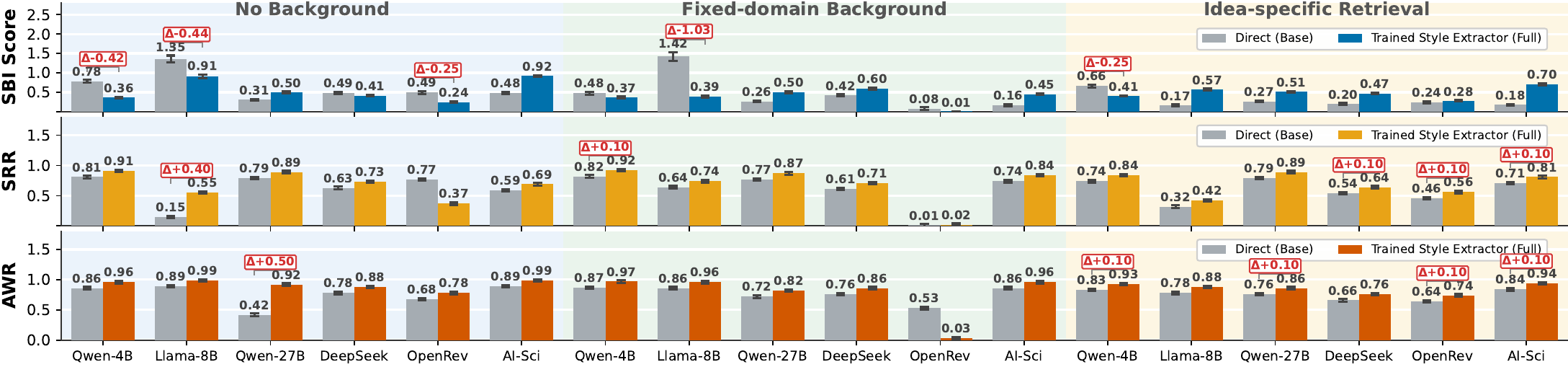}
}
\caption{
Comparison between direct judging and the trained SciStyleExtractor across
six judge models.
From left to right, the three column groups correspond to no background,
fixed-domain background, and idea-specific retrieval.
The top, middle, and bottom rows report SBI, SRR, and AWR, respectively;
lower SBI and higher SRR/AWR indicate better judging robustness.
}
\label{fig:variant-level-effects}
\end{figure*}

\subsection{Main Result}
We first evaluate whether LLM-as-a-Judge methods remain reliable under controlled style variations. We compare different judges across three background settings using SBI, SRR, and AWR, measuring style invariance, substance sensitivity, and adversarial robustness, respectively. Table~\ref{tab:method_comparison} summarizes the results. Direct judges show substantial style sensitivity and limited substance discrimination, whereas SciStyleExtractor achieves a better balance across the three objectives, although residual ranking instability remains.

(1) Direct judging is not robust.
Across four general-purpose judges and three settings, Direct Judge achieves
SBI/SRR/AWR scores of $0.566/0.504/0.554$, indicating substantial style
sensitivity and limited substance discrimination. Dimension-level analysis
shows that clarity ($1.185$) and feasibility ($0.702$) are most affected by
presentation style. Additional context does not consistently improve
robustness: retrieval reduces SBI but provides limited gains in SRR and AWR.
(2) Low SBI does not necessarily imply robustness.
Under fixed-domain background, Direct OpenReviewer obtains near-zero SBI but
also near-zero SRR and AWR, suggesting score collapse rather than effective
debiasing. With idea-specific retrieval and SciStyleExtractor, its scores
improve to $0.280/0.560/0.740$. These results highlight that SBI, SRR, and
AWR should be interpreted jointly.

(3) SciStyleExtractor achieves the best balance.
The trained SciStyleExtractor achieves $0.501/0.759/0.899$, outperforming
Direct Judge and Style-CoT ($0.581/0.551/0.571$). LLM-based style injection
improves AWR ($0.853$) but increases SBI to $0.999$, suggesting
over-correction. The extractor's gains mainly come from improved SRR and AWR
rather than SBI reduction, indicating that it improves substance sensitivity
while maintaining style robustness instead of simply flattening scores.
(4) Ranking instability remains.
The average absolute rank shifts are $2.116$, $2.613$, and $2.817$ for Direct
Judge, Style-CoT, and the KL-trained SciStyleExtractor, respectively. Rank
shifts are larger for substance/logic variants ($2.948$) and hybrid variants
($3.693$) than for rhetoric/framing variants ($1.940$). The larger shift of
SciStyleExtractor does not necessarily indicate worse robustness, as shifts
caused by substantive changes reflect desirable sensitivity, while
style-only shifts reveal residual bias. Hybrid variants remain the most
challenging due to the combination of degraded substance and persuasive
presentation.

\begin{figure}[t]
    \centering
    \includegraphics[width=\columnwidth]{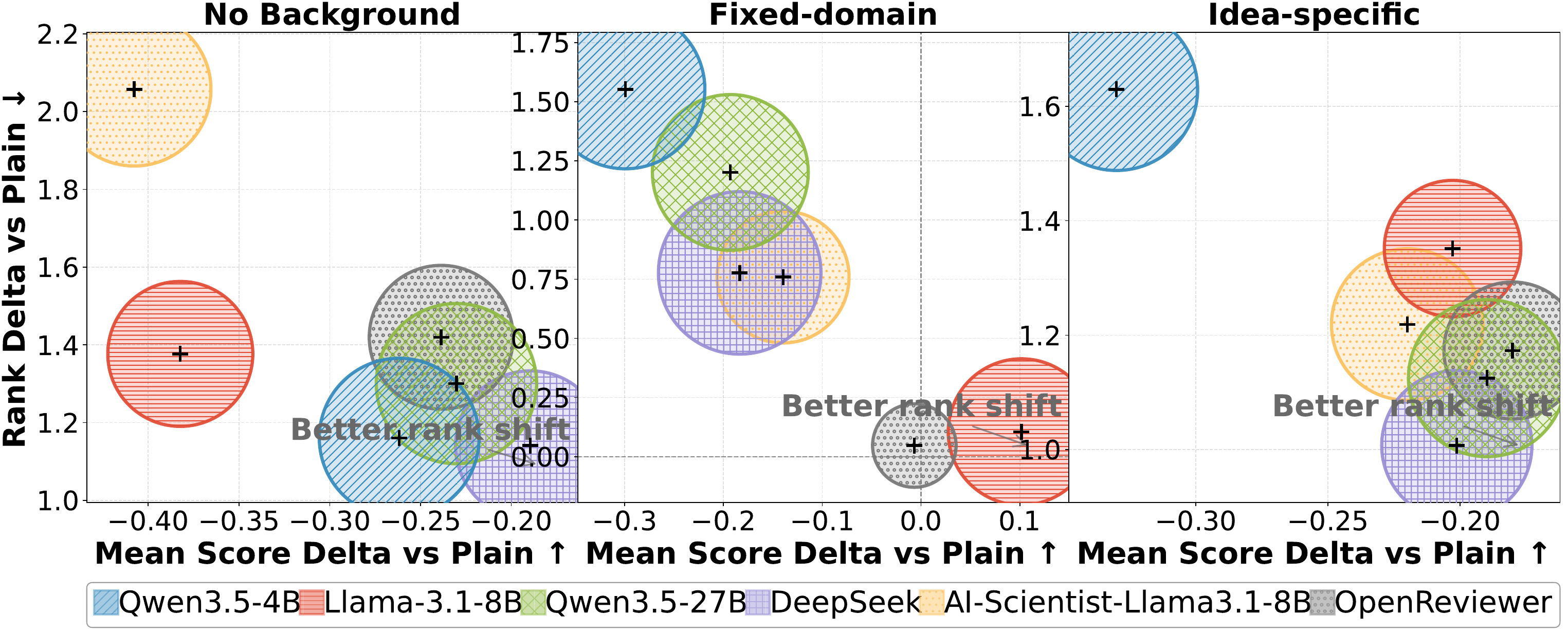}
    \caption{
    Score and rank shifts relative to the Plain reference. The x-axis denotes mean score delta and the y-axis
    denotes rank delta; lower rank shifts indicate more stable rankings.
    }
    \label{fig:score-rank-shift}
\end{figure}

\subsection{Additional Analysis}
\label{sec:variant-level-effects}
Variant-level Perturbation Effects.
To further unpack the main results, we examine how the effect of
SciStyleExtractor varies across judge models and background settings.
As shown in Fig.~\ref{fig:variant-level-effects}, we compare direct
judging with the trained SciStyleExtractor across six judge models and three
background settings. SciStyleExtractor generally improves SRR and AWR,
indicating stronger substance recognition and adversarial robustness, while
its effect on SBI varies across judges and settings. These results show that
the extractor provides a more balanced evaluation overall, but does not
uniformly eliminate style sensitivity.

Score and Rank Shifts.
We further analyze whether score changes translate into stable ranking
behavior. As shown in Fig.~\ref{fig:score-rank-shift}, the relationship
between mean score shifts and rank shifts varies substantially across judges
and background settings. Several judges exhibit noticeable ranking changes
even when their average score shifts are limited, showing that pointwise score
stability does not necessarily guarantee ranking stability. The results
further indicate that SciStyleExtractor improves overall robustness but does
not completely remove style-induced changes in relative ranking.

Top-$K$ Membership Changes.
For each style condition, we rank all 600 ideas and compare its Top-$K$ set
with the Plain reference. \textit{Drop out} and \textit{Enter} denote ideas
leaving and entering the Plain Top-$K$ after transformation. As shown in
Fig.~\ref{fig:topk-membership}, the membership-change rate rises from
$4.6\%$ at Top-5 to $15.9\%$ at Top-30, showing that style variation can alter
candidate selection and lead to downstream screening or resource-allocation
errors.

\subsection{Rewrite Validation}
\label{sec}
First, to ensure the stylistic accuracy of the generated variants and the stability of the evaluation process, we further conduct a style consistency evaluation. 
Specifically, we manually annotate the source ideas and their corresponding generated variants, comparing the human evaluation results against the outcomes of the LLM judges, as well as performing cross-comparisons among different LLM judges. 
The evaluation results indicate that the target style of the generated variants can be consistently identified by various LLM judges, and the automated evaluation results maintain a high degree of agreement with human assessments. 
Therefore, this consistency manifested not only across different LLM judges but also between the LLM judges and human evaluators demonstrates the effectiveness of our generated variants in terms of style control and evaluation reliability.

\begin{table}[htbp]
\centering
\renewcommand{\arraystretch}{0.65}
\resizebox{\columnwidth}{!}{%
\begin{tabular}{lccc}
\toprule
\textbf{Comparison Pair} & \textbf{Spearman (p)} & \textbf{Overall Top-50} & \textbf{Five-score Top-5} \\
\midrule
ChatGPT vs Human & 0.88 & 0.80 & 0.80 \\
DeepSeek-V3.2 vs Human & 0.84 & 0.78 & 0.40 \\
Kimi vs Human & 0.81 & 0.68 & 0.60 \\
ChatGPT vs DeepSeek-V3.2 & 0.82 & 0.90 & 0.74 \\
ChatGPT vs Kimi & 0.92 & 0.93 & 0.82 \\
DeepSeek-V3.2 vs Kimi & 0.90 & 0.94 & 0.76 \\

\bottomrule
\end{tabular}%
}
\label{tab:simulated_human_llm_pairwise_agreement}
\caption{Agreement Between Human and LLM Judges.}
\end{table}

Second, to validate that our perturbations modify presentation style while preserving scientific content, we randomly sampled 320 source-variant pairs for human evaluation. Two annotators assessed three aspects: (1) preservation of scientific substance, including the research question, mechanism, variables, constraints, methodology, and contribution; (2) consistency of perceived scientific quality; and (3) whether differences were primarily stylistic rather than content-related. Disagreements were resolved through adjudication without access to downstream evaluation results. As shown in Table~\ref{tab:style_validation}, 93.1\% of pairs preserved the original scientific substance, 90.9\% maintained equivalent perceived quality, and 97.9\% were identified as presentation-level variations. These results confirm that our perturbations primarily alter stylistic expression while preserving the underlying scientific value of the ideas.

\begin{table}[t]
    \centering
    \setlength{\tabcolsep}{1pt}
    \renewcommand{\arraystretch}{0.45}
    \begin{tabular}{lcccc}
    \toprule
    \textbf{Variant} & 
    \textbf{\#Pairs} &
    \textbf{Substance} &
    \textbf{Quality} &
    \textbf{Style} \\
    & &
    \textbf{Preserve (\%)} &
    \textbf{Same (\%)} &
    \textbf{Only (\%)}\\
    \midrule
    Paraphrase          & 40 & 97.5 & 95.0 & 100.0 \\
    Plain Core          & 40 & 95.0 & 92.5 & 98.3 \\
    Verbose             & 40 & 92.5 & 90.0 & 98.3 \\
    Grand Narrative     & 40 & 90.0 & 87.5 & 96.7 \\
    Overconfident       & 40 & 92.5 & 90.0 & 98.3 \\
    Ultimate Hype       & 40 & 87.5 & 82.5 & 95.0 \\
    \midrule
    \textbf{Overall}    
    & \textbf{240} 
    & \textbf{92.5} 
    & \textbf{89.7} 
    & \textbf{97.8} \\
    \bottomrule
    \end{tabular}
    \caption{
    Human validation of style-only transformations.
    Experts evaluate whether transformed variants preserve scientific substance,
    maintain perceived scientific quality, and differ primarily in presentation style.
    }
    \label{tab:style_validation}
\end{table}

\subsection{Ablation}
\label{sec:ablation}
We fix Qwen3.5-4B as the target judge and compare five configurations:
direct judging, Style-CoT, LLM-based style injection, and SciStyleExtractor
trained with either SFT or SFT+KL. This comparison examines whether robustness
gains come from generic prompting, externally generated style signals, or a
learned extractor. Qwen3.5-27B provides teacher supervision, while the
Qwen3.5-4B extractor is implemented as a LoRA adapter and trained on 2,000 training pairs for two epochs. For SFT and KL, the extractor is
initialized with SFT and further aligned to the teacher's output distribution
through KL distillation. Results are averaged across the three background
settings. As shown in Table.~\ref{tab:ablation_robustness}, Style-CoT performs
worse than direct judging, indicating that simply prompting the judge to
consider style is insufficient. LLM-based injection achieves a high AWR but
substantially worsens SBI and SRR, suggesting over-correction caused by noisy
or overly strong style signals. In contrast, both trained extractors improve
the overall balance among the three metrics: SFT+KL achieves the lowest SBI,
while SFT-only obtains slightly higher SRR and AWR. These results suggest that
KL distillation mainly strengthens style invariance, whereas SFT better
preserves substance discrimination and adversarial robustness.
\begin{table}[t]
\centering
\small
\setlength{\tabcolsep}{4pt}
\renewcommand{\arraystretch}{0.7}
\resizebox{\columnwidth}{!}{
\begin{tabular}{lccc}
\hline
\textbf{Style Signal Source}
& \textbf{SBI $\downarrow$}
& \textbf{SRR $\uparrow$}
& \textbf{AWR $\uparrow$} \\
\hline
None / Direct Judge & 0.639 & 0.800 & 0.900 \\
Style-CoT Prompt & 0.746 & 0.750 & 0.773 \\
LLM-based Style Injection & 1.279 & 0.497 & 1.000 \\
\hline
Qwen3.5-4B Style Extractor (KL) & 0.380 & 0.890 & 0.953 \\
Qwen3.5-4B Style Extractor (SFT) & 0.406 & 0.893 & 0.973 \\
\hline
\end{tabular}
}
\caption{
Ablation of style-signal sources.
We compare direct judging, style-aware prompting, LLM-based injection, and
trained style extractors. SBI is averaged across three evaluation settings
and style-only variants.
}
\label{tab:ablation_robustness}
\end{table}

\section{Conclusion}
This work introduces SciStyleBench to diagnose and mitigate stylistic bias in
scientific idea evaluation. Experiments across diverse judges and background
settings show that direct LLM-as-a-Judge methods remain sensitive to
presentation style and struggle to reliably distinguish substantive scientific
quality. We further show that low style sensitivity alone is insufficient, as
low SBI may result from score collapse rather than genuine robustness.
SciStyleExtractor improves the balance among style invariance, substance
recognition, and adversarial robustness, while residual ranking instability
remains. Overall, robust scientific idea evaluation requires judges to
suppress style-related nuisance signals while preserving sensitivity to
scientific substance, logical validity, and practical feasibility.

\clearpage
\bibliography{aaai2027}

@inproceedings{liu2026researchbench,
  title={Researchbench: Benchmarking llms in scientific discovery via inspiration-based task decomposition},
  author={Liu, Yujie and Yang, Zonglin and Xie, Tong and Ni, Jinjie and Gao, Ben and Li, Yuqiang and Tang, Shixiang and Ouyang, Wanli and Cambria, Erik and Zhou, Dongzhan},
  booktitle={Findings of the Association for Computational Linguistics: ACL 2026},
  pages={13187--13207},
  year={2026}
}

@dataset{paperzilla_rag_retrieval_2024,
  title={Paperzilla RAG Retrieval Benchmark: Multi-Annotator Dataset for Scientific Paper Retrieval},
  author={Paperzilla Team},
  year={2024},
  publisher={HuggingFace},
  url={https://huggingface.co/datasets/paperzilla/paperzilla-rag-retrieval-250}
}

@article{zheng2023judging,
  title={Judging llm-as-a-judge with mt-bench and chatbot arena},
  author={Zheng, Lianmin and Chiang, Wei-Lin and Sheng, Ying and Zhuang, Siyuan and Wu, Zhanghao and Zhuang, Yonghao and Lin, Zi and Li, Zhuohan and Li, Dacheng and Xing, Eric and others},
  journal={Advances in neural information processing systems},
  volume={36},
  pages={46595--46623},
  year={2023}
}

@inproceedings{wang2024large,
  title={Large language models are not fair evaluators},
  author={Wang, Peiyi and Li, Lei and Chen, Liang and Cai, Zefan and Zhu, Dawei and Lin, Binghuai and Cao, Yunbo and Kong, Lingpeng and Liu, Qi and Liu, Tianyu and others},
  booktitle={Proceedings of the 62nd Annual Meeting of the Association for Computational Linguistics (Volume 1: Long Papers)},
  pages={9440--9450},
  year={2024}
}

@article{wang2024scipip,
title={SciPIP: An LLM-based Scientific Paper Idea Proposer},
author={Wang, Wenhai and Gu, Lei and Zhang, Lei and Luo, Yuxiao and Dai, Yutong and Shen, Chao and Xie, Lei and Lin, Boyan and He, Xiaodong and Ye, Jiebo},
journal={arXiv preprint arXiv:2410.23166},
year={2024}
}

@article{xiong2024kgcoi,
title={Improving Scientific Hypothesis Generation with Knowledge Grounded Large Language Models},
author={Xiong, Guangyi and Xie, Enze and Shariatmadari, Alireza H. and Guo, Sheng and Bekiranov, Stefan and Zhang, Aidong},
journal={arXiv preprint arXiv:2411.02382},
year={2024}
}

@article{radensky2024scideator,
title={Scideator: Human-LLM Compound System for Scientific Ideation through Facet Recombination and Novelty Evaluation},
author={Radensky, Marisa and Shahid, Salman and Fok, Raymond and Siangliulue, Pao and Hope, Tom and Weld, Daniel S.},
journal={arXiv preprint arXiv:2409.14634},
year={2024}
}

@inproceedings{pu2025ideasynth,
title={IdeaSynth: Iterative Research Idea Development Through Evolving and Composing Idea Facets with Literature-Grounded Feedback},
author={Pu, Kevin and Feng, Kevin J. K. and Grossman, Tovi and Hope, Tom and Dalvi Mishra, Bhavana and Latzke, Matt and Bragg, Jonathan and Chang, Joseph Chee and Siangliulue, Pao},
booktitle={Proceedings of the CHI Conference on Human Factors in Computing Systems},
year={2025}
}

@inproceedings{liu2025personaflow,
title={PersonaFlow: Designing LLM-Simulated Expert Perspectives for Enhanced Research Ideation},
author={Liu, Yang and Sharma, Pratyush and Oswal, Mihir and Xia, Haijun and Huang, Yun},
booktitle={Proceedings of the ACM Conference},
year={2025}
}

@inproceedings{baek2025researchagent,
title={ResearchAgent: Iterative Research Idea Generation over Scientific Literature with Large Language Models},
author={Baek, Jinheon and Jauhar, Sujay Kumar and Cucerzan, Silviu and Hwang, Seung-won J.},
booktitle={Proceedings of NAACL 2025},
year={2025}
}

@inproceedings{wang2024scimon,
  title={SciMON: Scientific Inspiration Machines Optimized for Novelty},
  author={Wang, Qingyun and Downey, Doug and Ji, Heng and Hope, Tom},
  booktitle={Proceedings of the 62nd Annual Meeting of the Association for Computational Linguistics (Volume 1: Long Papers)},
  pages={279--299},
  year={2024},
  address={Bangkok, Thailand},
  publisher={Association for Computational Linguistics},
  doi={10.18653/v1/2024.acl-long.18}
}

@misc{radensky2026humanllmcompoundscientificideation,
      title={Human-LLM Compound System for Scientific Ideation through Facet Recombination and Novelty Evaluation}, 
      author={Marissa Radensky and Simra Shahid and Raymond Fok and Pao Siangliulue and Tom Hope and Daniel S. Weld},
      year={2026},
      eprint={2409.14634},
      archivePrefix={arXiv},
      primaryClass={cs.HC},
      url={https://arxiv.org/abs/2409.14634}, 
}

@misc{luo2025llm4srsurveylargelanguage,
      title={LLM4SR: A Survey on Large Language Models for Scientific Research}, 
      author={Ziming Luo and Zonglin Yang and Zexin Xu and Wei Yang and Xinya Du},
      year={2025},
      eprint={2501.04306},
      archivePrefix={arXiv},
      primaryClass={cs.CL},
      url={https://arxiv.org/abs/2501.04306}, 
}

@misc{guo2024ideabenchbenchmarkinglargelanguage,
      title={IdeaBench: Benchmarking Large Language Models for Research Idea Generation}, 
      author={Sikun Guo and Amir Hassan Shariatmadari and Guangzhi Xiong and Albert Huang and Eric Xie and Stefan Bekiranov and Aidong Zhang},
      year={2024},
      eprint={2411.02429},
      archivePrefix={arXiv},
      primaryClass={cs.CL},
      url={https://arxiv.org/abs/2411.02429}, 
}

@misc{ruan2026evaluatingllmsdivergentthinking,
      title={Evaluating LLMs' Divergent Thinking Capabilities for Scientific Idea Generation with Minimal Context}, 
      author={Kai Ruan and Xuan Wang and Jixiang Hong and Peng Wang and Yang Liu and Hao Sun},
      year={2026},
      eprint={2412.17596},
      archivePrefix={arXiv},
      primaryClass={cs.CL},
      url={https://arxiv.org/abs/2412.17596}, 
}

@misc{lu2024aiscientistfullyautomated,
      title={The AI Scientist: Towards Fully Automated Open-Ended Scientific Discovery}, 
      author={Chris Lu and Cong Lu and Robert Tjarko Lange and Jakob Foerster and Jeff Clune and David Ha},
      year={2024},
      eprint={2408.06292},
      archivePrefix={arXiv},
      primaryClass={cs.AI},
      url={https://arxiv.org/abs/2408.06292}, 
}

@article{Gottweis_2026,
  title   = {Accelerating scientific discovery with Co-Scientist},
 author = {Gottweis, Juraj and Weng, Wei-Hung and Daryin, Alexander and et al.},
 journal = {Nature},
  year    = {2026},
  DOI     = {10.1038/s41586-026-10644-y}
}

@article{gelles2024resourcedemocratization,
  title={Resource Democratization: Is Compute the Binding Constraint on AI Research?},
  author={Gelles, Rebecca and Kinoshita, Veronica and Musser, Micah and Dunham, James},
  journal={Proceedings of the AAAI Conference on Artificial Intelligence},
  year={2024},
  volume={38},
  number={18},
  pages={19840--19848},
  doi={10.1609/aaai.v38i18.29959},
  publisher={Association for the Advancement of Artificial Intelligence (AAAI)},
  issn={2159-5399},
  month={Mar},
  url={http://dx.doi.org/10.1609/aaai.v38i18.29959},
}

@misc{qiu2025aiideabench2025,
      title={AI Idea Bench 2025: AI Research Idea Generation Benchmark}, 
      author={Yansheng Qiu and Haoquan Zhang and Zhaopan Xu and Ming Li and Diping Song and Zheng Wang and Kaipeng Zhang},
      year={2025},
      eprint={2504.14191},
      archivePrefix={arXiv},
      primaryClass={cs.AI},
      url={https://arxiv.org/abs/2504.14191}, 
}

@misc{li2026graph2idearetrievalaugmentedscientificideageneration,
      title={Graph2Idea:Retrieval-Augmented Scientific Idea Generation with Graph-Structured Contexts}, 
      author={Xu Li and Hanzhe Tu and Xun Han},
      year={2026},
      eprint={2606.09105},
      archivePrefix={arXiv},
      primaryClass={cs.AI},
      url={https://arxiv.org/abs/2606.09105}, 
}

@misc{dong2026evolvingideagraphslearnable,
      title={Evolving Idea Graphs with Learnable Edits-and-Commits for Multi-Agent Scientific Ideation}, 
      author={Jiangwen Dong and Bo Li and Wanyu Lin},
      year={2026},
      eprint={2605.04922},
      archivePrefix={arXiv},
      primaryClass={cs.MA},
      url={https://arxiv.org/abs/2605.04922}, 
}

@misc{su2025headsbetteroneimproved,
      title={Many Heads Are Better Than One: Improved Scientific Idea Generation by A LLM-Based Multi-Agent System}, 
      author={Haoyang Su and Renqi Chen and Shixiang Tang and Zhenfei Yin and Xinzhe Zheng and Jinzhe Li and Biqing Qi and Qi Wu and Hui Li and Wanli Ouyang and Philip Torr and Bowen Zhou and Nanqing Dong},
      year={2025},
      eprint={2410.09403},
      archivePrefix={arXiv},
      primaryClass={cs.AI},
      url={https://arxiv.org/abs/2410.09403}, 
}

@misc{keya2025sciideacontextawarescientificideation,
      title={SCI-IDEA: Context-Aware Scientific Ideation Using Token and Sentence Embeddings}, 
      author={Farhana Keya and Gollam Rabby and Prasenjit Mitra and Sahar Vahdati and Sören Auer and Yaser Jaradeh},
      year={2025},
      eprint={2503.19257},
      archivePrefix={arXiv},
      primaryClass={cs.CL},
      url={https://arxiv.org/abs/2503.19257}, 
}

@article{shahhosseini2025large,
  title={Large Language Models for Scientific Idea Generation: A Creativity-Centered Survey},
  author={Shahhosseini, Fatemeh and Marioriyad, Arash and Momen, Ali and Baghshah, Mahdieh Soleymani and Rohban, Mohammad Hossein and Javanmard, Shaghayegh Haghjooy},
  journal={arXiv preprint arXiv:2511.07448},
  year={2025}
}

@misc{si2024llmsgeneratenovelresearch,
      title={Can LLMs Generate Novel Research Ideas? A Large-Scale Human Study with 100+ NLP Researchers}, 
      author={Chenglei Si and Diyi Yang and Tatsunori Hashimoto},
      year={2024},
      eprint={2409.04109},
      archivePrefix={arXiv},
      primaryClass={cs.CL},
      url={https://arxiv.org/abs/2409.04109}, 
}

@article{chen2026mlr,
  title={Mlr-bench: Evaluating ai agents on open-ended machine learning research},
  author={Chen, Hui and Xiong, Miao and Lu, Yujie and Han, Wei and Deng, Ailin and He, Yufei and Wu, Jiaying and Li, Yibo and Liu, Yue and Hooi, Bryan},
  journal={Advances in Neural Information Processing Systems},
  volume={38},
  year={2026}
}

@article{si2025ideation,
  title={The ideation-execution gap: Execution outcomes of llm-generated versus human research ideas},
  author={Si, Chenglei and Hashimoto, Tatsunori and Yang, Diyi},
  journal={arXiv preprint arXiv:2506.20803},
  year={2025}
}

@article{kulkarni2025scientific,
  title={Scientific hypothesis generation and validation: Methods, datasets, and future directions},
  author={Kulkarni, Adithya and Alotaibi, Fatimah and Zeng, Xinyue and Wu, Longfeng and Zeng, Tong and Yao, Barry Menglong and Liu, Minqian and Zhang, Shuaicheng and Huang, Lifu and Zhou, Dawei},
  journal={arXiv preprint arXiv:2505.04651},
  year={2025}
}

@article{liu2025hypobench,
  title={Hypobench: Towards systematic and principled benchmarking for hypothesis generation},
  author={Liu, Haokun and Huang, Sicong and Hu, Jingyu and Zhou, Yangqiaoyu and Tan, Chenhao},
  journal={arXiv preprint arXiv:2504.11524},
  year={2025}
}

@article{kon2025exp,
  title={Exp-bench: Can ai conduct ai research experiments?},
  author={Kon, Patrick Tser Jern and Liu, Jiachen and Zhu, Xinyi and Ding, Qiuyi and Peng, Jingjia and Xing, Jiarong and Huang, Yibo and Qiu, Yiming and Srinivasa, Jayanth and Lee, Myungjin and others},
  journal={arXiv preprint arXiv:2505.24785},
  year={2025}
}

@article{jie2026capability,
  title={Capability-Aware Early-Stage Research Idea Evaluation},
  author={Jie, Renlong and Chu, Chen and Wang, Zhen},
  journal={arXiv preprint arXiv:2601.12473},
  year={2026}
}

@article{liu2026whoowns,
  title={Who Owns Creativity and Who Does the Work? Trade-offs in LLM-Supported Research Ideation},
  author={Houjiang Liu and Yujin Choi and Sanjana Gautam and Gabriel Jaffe and Soo Young Rieh and Matthew Lease},
  journal={ArXiv.org},
  year={2026},
}

@article{nigam2024accelerontool,
  title={Acceleron: A Tool to Accelerate Research Ideation},
  author={Harshit Nigam and Manasi Patwardhan and Lovekesh Vig and Gautam Shroff},
  journal={arXiv (Cornell University)},
  year={2024},
  doi={10.48550/arxiv.2403.04382},
}

@article{sinhahajari2026limits,
  title={On the Limits of LLM-as-Judge for Scientific Novelty Assessment},
  author={Sinhahajari, Soumitra and Majumder, Navonil and Poria, Soujanya},
  journal={arXiv preprint arXiv:2606.12071},
  year={2026}
}

@article{li2025automated,
  title={Automated creativity evaluation for large language models: A reference-based approach},
  author={Li, Ruizhe and Zhu, Chiwei and Xu, Benfeng and Wang, Xiaorui and Mao, Zhendong},
  journal={arXiv preprint arXiv:2504.15784},
  year={2025}
}

@article{rabeyah2024llms,
  title={Do LLMs Agree on the Creativity Evaluation of Alternative Uses?},
  author={Rabeyah, Abdullah Al and G{\'o}es, Fabr{\'\i}cio and Volpe, Marco and Medeiros, Talles},
  journal={arXiv preprint arXiv:2411.15560},
  year={2024}
}

@inproceedings{afzal2026beyond,
  title={Beyond" not novel enough": Enriching scholarly critique with llm-assisted feedback},
  author={Afzal, Osama Mohammed and Nakov, Preslav and Hope, Tom and Gurevych, Iryna},
  booktitle={Proceedings of the 19th Conference of the European Chapter of the Association for Computational Linguistics (Volume 1: Long Papers)},
  pages={2648--2671},
  year={2026}
}

@article{li2024llms,
  title={Llms-as-judges: a comprehensive survey on llm-based evaluation methods},
  author={Li, Haitao and Dong, Qian and Chen, Junjie and Su, Huixue and Zhou, Yujia and Ai, Qingyao and Ye, Ziyi and Liu, Yiqun},
  journal={arXiv preprint arXiv:2412.05579},
  year={2024}
}

@article{shen2026navigating,
  title={Navigating Ideation Space: Decomposed Conceptual Representations for Positioning Scientific Ideas},
  author={Shen, Yuexi and Liu, Minqian and Zhou, Dawei and Huang, Lifu},
  journal={arXiv preprint arXiv:2601.08901},
  year={2026}
}

@inproceedings{koo2026auditing,
  title={Auditing the Judge: Human-Grounded Bias Discovery, Quantification, and Mitigation in LLM Judges},
  author={Koo, Hamin and Jung, ChanJoo and Wu, Fangzhao and Kim, Jaehyung},
  booktitle={Trustworthy AI for Good (AI4GOOD) Workshop@ ICML 2026}
}

@article{cao2025evaluating,
  title={Evaluating text creativity across diverse domains: A dataset and large language model evaluator},
  author={Cao, Qian and Wang, Xiting and Yuan, Yuzhuo and Liu, Yahui and Luo, Fang and Song, Ruihua},
  journal={arXiv preprint arXiv:2505.19236},
  year={2025}
}

@article{rasheed2026fluent,
  title={From fluent to verifiable: Claim-level auditability for deep research agents},
  author={Rasheed, Razeen A and Banerjee, Somnath and Mukherjee, Animesh and Hazra, Rima},
  journal={arXiv preprint arXiv:2602.13855},
  year={2026}
}

@article{wang2026firstresearch,
  title={FirstResearch: Auditable Question Formation for LLM Scientific Discovery Agents},
  author={Wang, Yufeng},
  journal={arXiv preprint arXiv:2607.05682},
  year={2026}
}

@article{ho2026soundnessbench,
  title={SoundnessBench: Can Your AI Scientist Really Tell Good Research Ideas from Bad Ones?},
  author={Ho, Sy-Tuyen and Liu, Minghui and Nghiem, Huy and Huang, Furong},
  journal={arXiv preprint arXiv:2605.30329},
  year={2026}
}

@article{ye2026proof,
  title={Proof of Time: A Benchmark for Evaluating Scientific Idea Judgments},
  author={Ye, Bingyang and Chen, Shan and Tu, Jingxuan and Liu, Chen and Xiong, Zidi and Schmidgall, Samuel and Bitterman, Danielle S},
  journal={arXiv preprint arXiv:2601.07606},
  year={2026}
}

@inproceedings{wu2025style,
  title={Style over substance: Evaluation biases for large language models},
  author={Wu, Minghao and Aji, Alham Fikri},
  booktitle={Proceedings of the 31st International Conference on Computational Linguistics},
  pages={297--312},
  year={2025}
}

@inproceedings{zhou2024mitigating,
  title={Mitigating the bias of large language model evaluation},
  author={Zhou, Hongli and Huang, Hui and Long, Yunfei and Xu, Bing and Zhu, Conghui and Cao, Hailong and Yang, Muyun and Zhao, Tiejun},
  booktitle={Proceedings of the 23rd Chinese National Conference on Computational Linguistics (Volume 1: Main Conference)},
  pages={1310--1319},
  year={2024}
}

@article{zhou2026toward,
  title={Toward robust LLM-based judges: taxonomic bias evaluation and debiasing optimization},
  author={Zhou, Hongli and Huang, Hui and Zhang, Rui and Chen, Kehai and Xu, Bing and Zhu, Conghui and Zhao, Tiejun and Yang, Muyun},
  journal={arXiv preprint arXiv:2603.08091},
  year={2026}
}

@article{yang2026any,
  title={Any large language model can be a reliable judge: Debiasing with a reasoning-based bias detector},
  author={Yang, Haoyan and Bao, Runxue and Xiao, Cao Danica and Ma, Jun and Bhatia, Parminder and Gao, Shangqian and Kass-Hout, Taha},
  journal={Advances in Neural Information Processing Systems},
  volume={38},
  pages={6318--6362},
  year={2026}
}

@inproceedings{zeng2024evaluating,
  title={Evaluating large language models at evaluating instruction following},
  author={Zeng, Zhiyuan and Yu, Jiatong and Gao, Tianyu and Meng, Yu and Goyal, Tanya and Chen, Danqi},
  booktitle={International Conference on Learning Representations},
  volume={2024},
  pages={40193--40219},
  year={2024}
}

@inproceedings{idahl2025openreviewer,
  title={Openreviewer: A specialized large language model for generating critical scientific paper reviews},
  author={Idahl, Maximilian and Ahmadi, Zahra},
  booktitle={Proceedings of the 2025 Conference of the Nations of the Americas Chapter of the Association for Computational Linguistics: Human Language Technologies (System Demonstrations)},
  pages={550--562},
  year={2025}
}

@article{bao2026contemporary,
  title={Contemporary AI lacks the imagination to diverge or negate in science},
  author={Bao, Honglin and Wu, Siyang and Liu, Xiao and Li, Sida and Cao, Shiyun and Evans, James A},
  journal={arXiv preprint arXiv:2606.08251},
  year={2026}
}

@inproceedings{ji2026finestate,
  title={FineState-Bench: Benchmarking State-Conditioned Grounding for Fine-grained GUI State Setting},
  author={Ji, Fengxian and Yang, Jingpu and Song, Zirui and Wang, Yuanxi and Cui, Zhexuan and Li, Yuke and Jiang, Qian and Chen, Xiuying},
  booktitle={Findings of the Association for Computational Linguistics: ACL 2026},
  pages={43073--43088},
  year={2026}
}

@inproceedings{ji2026servimage,
  title={ServImage: An image generation and editing benchmark from real-world commercial imaging services},
  author={Ji, Fengxian and Yang, Jingpu and Song, Zirui and Gao, Lang and Liang, Junhong and Chen, Zhenhao and Zhang, Jinghui and Chen, Xiuying},
  booktitle={Proceedings of the 64th Annual Meeting of the Association for Computational Linguistics (Volume 1: Long Papers)},
  pages={43504--43529},
  year={2026}
}

@article{yang2026labguard,
  title={LabGuard: Grounding Natural-Language Laboratory Rules into Runtime Guards for Embodied Laboratory Agents},
  author={Yang, Jingpu and Ji, Fengxian and Lai, Zhengzhao and Cui, Zhexuan and Ouyang, Guangxian and Jiang, Qian and Zhang, Fan and Peng, Min and Xie, Qianqian and Nakov, Preslav and others},
  journal={arXiv preprint arXiv:2606.31045},
  year={2026}
}

@article{ji2026parametric,
  title={Parametric Memory Decoding for Zero-Shot Routing in LoRA-Based External Parametric Memory},
  author={Ji, Fengxian and Xie, Zhuohan and Yang, Jingpu and Zhang, Fan and Song, Zirui and Chen, Xiuying},
  journal={arXiv preprint arXiv:2607.04118},
  year={2026}
}

\end{document}